\documentclass[letterpaper, 10 pt, conference]{ieeeconf}  

\IEEEoverridecommandlockouts                              

\usepackage{amsmath} 
\usepackage{amssymb}  
\usepackage{amsfonts}  
\usepackage{colortbl}	
\usepackage{multirow}	
\usepackage[usenames,dvipsnames,table]{xcolor}
\usepackage{array}
\usepackage{colortbl}	
\usepackage{url}
\usepackage{flushend} 

\usepackage{graphicx}
\usepackage[percent]{overpic} 
\usepackage{subcaption}
\usepackage{bm}

\newcommand{\regressor}{\ensuremath{\Phi(\mathbf{x})}}
\newcommand{\regressorNum}[1]{\ensuremath{\Phi(\mathbf{x}_{#1})}}
\newcommand{\outputQuant}{\ensuremath{\mathbf{y}}}
\newcommand{\nrOfSamples}{n}
\newcommand{\inertialParameters}{\bm{\pi}}
\newcommand{\estimatedInertialParameters}{\hat{\bm{\pi}}}
\newcommand{\RLSinputMat}{Z}
\newcommand{\RLSoutputMat}{U}
\newcommand{\RLSinputSample}{\mathbf{z}}
\newcommand{\RLSoutputSample}{\mathbf{u}}
\DeclareMathOperator*{\argmin}{\arg\!\min}

\title{\LARGE \bf
Incremental Semiparametric Inverse Dynamics Learning
}

\author{Raffaello Camoriano$^{*\dagger}$, Silvio Traversaro$^{\ddagger}$,\\ Lorenzo Rosasco$^{\diamond}$, Giorgio Metta$^{\vartriangle}$, and Francesco Nori$^{\ddagger}$
\thanks{$^{*}$Corresponding author.}
\thanks{$^{\dagger}$Raffaello Camoriano is with iCub Facility, Istituto Italiano di Tecnologia, Via Morego 30, Genoa 16163, Italy, and DIBRIS, Universit\`{a} degli Studi di Genova, Via All'Opera Pia, 13, Genoa 16145, Italy. Email: {\tt\small raffaello.camoriano@iit.it}}%
\thanks{$^{\ddagger}$Silvio Traversaro and Francesco Nori are with RBCS Department, Istituto Italiano di Tecnologia, Via Morego 30, Genoa 16163, Italy. Email: {\tt\small name.surname@iit.it}}%
\thanks{$^{\diamond}$Lorenzo Rosasco is with LCSL, Istituto Italiano di Tecnologia and Massachusetts Institute of Technology, Cambridge, MA 02139, USA, and DIBRIS, Universit\`{a} degli Studi di Genova, Via All'Opera Pia, 13, Genoa 16145, Italy. Email: {\tt\small lrosasco@mit.edu}}%
\thanks{$^{\vartriangle}$Giorgio Metta is with iCub Facility, Istituto Italiano di Tecnologia, Via Morego 30, Genoa 16163, Italy. Email: {\tt\small giorgio.metta@iit.it}}
}

\begin{document}

\maketitle
\thispagestyle{empty}
\pagestyle{empty}

\begin{abstract}
This paper presents a novel approach for incremental semiparametric inverse dynamics learning. 
In particular, we consider the mixture of two approaches: Parametric modeling based on rigid body dynamics equations 
and nonparametric modeling based on incremental kernel methods, with no prior information on the mechanical properties 
of the system. This yields to an incremental semiparametric approach, leveraging the advantages of both the parametric and 
nonparametric models. We validate the proposed technique learning the dynamics of one arm of the iCub humanoid robot. 
\end{abstract}

\section{INTRODUCTION}

In order to control a robot a model describing the relation between the actuator inputs, 
the interactions with the world and bodies accelerations is required. This model is called the \emph{dynamics} model of the robot. 
A dynamics model can be obtained from first principles in mechanics, using the techniques of rigid body dynamics (RBD) \cite{reference/robo/FeatherstoneO08}, 
resulting in a \emph{parametric model} in which the values of physically meaningful parameters 
must be provided to complete the fixed structure of the model. 
Alternatively, the dynamical model can be obtained from experimental data using Machine Learning techniques, resulting in a \emph{nonparametric model}.

Traditional dynamics parametric methods are based on several assumptions, such as rigidity of links or that friction has
a simple analytical form, which may not be accurate in real systems.  
On the other hand, nonparametric methods based on algorithms such as  Kernel Ridge Regression (KRR) \cite{hoerl1970ridgeregression,SaundersGV98,cristianini2000introduction}, Kernel Regularized Least Squares (KRLS) \cite{rifkin2003regularized} or Gaussian Processes \cite{rasmussen2006}
can model dynamics by extrapolating the input-output relationship directly from the available data\footnote{Note that KRR and KRLS have a very similar formulation, and that these are also equivalent to the techniques derived from Gaussian Processes, as explained for instance in Chapter 6 of  \cite{cristianini2000introduction}.}. 
If a suitable kernel function is chosen, then the nonparametric model is a universal approximator 
which can account for the dynamics effects which are not considered by the parametric model. 
Still, nonparametric models have no prior knowledge about the target function to be approximated. Therefore, they need a sufficient amount of training examples in order to produce accurate predictions on the entire input space.
If the learning phase has been performed offline, both approaches are sensitive to the variation of the mechanical properties over long time spans, which are mainly caused by temperature shifts and wear. Even the inertial parameters can change over time. For example if the robot grasps a heavy object, the resulting change in dynamics can be described by a change of the inertial parameters of the hand.  
A solution to this problem is to address the variations of the identified system properties by learning \emph{incrementally}, continuously
updating the model as long as new data becomes available. 
In this paper we propose a novel technique that joins parametric and nonparametric model learning in an incremental fashion.

\begin{table}[t] 
\caption{Summary of related works on semiparametric or incremental robot dynamics learning.}
\begin{center}
\resizebox{0.48\textwidth}{!}{%
\begin{tabular}{| c | c c |} 
\hline
         \rowcolor[gray]{.9}                                                                                                        
          \textbf{Author, Year}                                   & \textbf{Parametric}     & \textbf{Nonparametric}  \\ 
          [0.5ex] \hline
         Nguyen-Tuong, 2010 \cite{Nguyen-TuongP10}      & Batch  &  Batch   \\
         \rowcolor[gray]{.9}                                                                                                        
         Gijsberts, 2011 \cite{GijsbertsM11}            &  -      &  Incremental   \\
         Tingfan Wu, 2012 \cite{conf/iros/wu12}         &  Batch &  Batch   \\
         \rowcolor[gray]{.9}                                                                                                        
         De La Cruz, 2012 \cite{conf/ifac/delacruz12}   &  CAD$^*$  &  Incremental  \\
         Camoriano, 2015                                &  Incremental &  Incremental \\
         \hline
\end{tabular} 
}
\end{center}

\label{table:soa} 
$^*$ In \cite{conf/ifac/delacruz12} the parametric part is used only for initializing the nonparametric model. 
\end{table}


Classical methods for physics-based dynamics modeling can be found in \cite{reference/robo/FeatherstoneO08}. 
These methods require to identify the mechanical parameters of the rigid 
bodies composing the robot \cite{conf/humanoids/Yamane11,conf/humanoids/TraversaroPMNN13,conf/humanoids/OgawaVO14,hollerbach2008model}, 
which can then be employed in model-based control and state estimation schemes.

 In \cite{Nguyen-TuongP10} the authors present a learning  technique which 
 combines prior knowledge about the physical structure of the mechanical
 system and learning from available data with Gaussian Process Regression (GPR) \cite{rasmussen2006}. 
 A similar approach is presented in \cite{conf/iros/wu12}. Both techniques 
  require an offline training phase and are not incremental, limiting them to scenarios in which the properties of the system do not change significantly over time. 
 
In \cite{conf/ifac/delacruz12} an incremental semiparametric robot dynamics
learning scheme based on Locally Weighted Projection Regression (LWPR) \cite{conf/icml/VijayakumarS00} is presented, that is initialized using a linearized parametric model. However, this approach uses a fixed parametric model, 
that is not updated as new data becomes available. Moreover, LWPR has been shown to underperform with respect to other methods (e.g. \cite{GijsbertsM11}).


In \cite{GijsbertsM11}, a fully nonparametric incremental approach for inverse dynamics learning with constant
 update complexity is presented, based on kernel methods \cite{schlkopf2002learning} (in particular KRR) and random features \cite{RahimiR07}. The incremental nature of this 
 approach allows for adaptation to changing conditions in time. The authors also show that the proposed algorithm outperforms other methods such as LWPR, GPR and Local Gaussian Processes (LGP) \cite{lgpr}, both in terms of accuracy and prediction time. Nevertheless, the fully nonparametric nature of this approach undermines the interpretability of the inverse dynamics model.


In this work we propose a method that is incremental with fixed update complexity (as \cite{GijsbertsM11}) and 
semiparametric (as  \cite{Nguyen-TuongP10} and \cite{conf/iros/wu12}). The fixed update complexity and prediction time are key properties of our method, enabling real-time performances. Both the parametric and nonparametric parts can be updated, 
as opposed to \cite{conf/ifac/delacruz12} in which only the nonparametric part is. A comparison between the existing literature and our incremental
method is reported in Table \ref{table:soa}.
We validate the proposed method with experiments performed on an arm of the iCub humanoid robot \cite{Metta2010}.
\begin{figure}[ht!]
\vspace{3mm}
\centering
\includegraphics[width=0.95\linewidth]{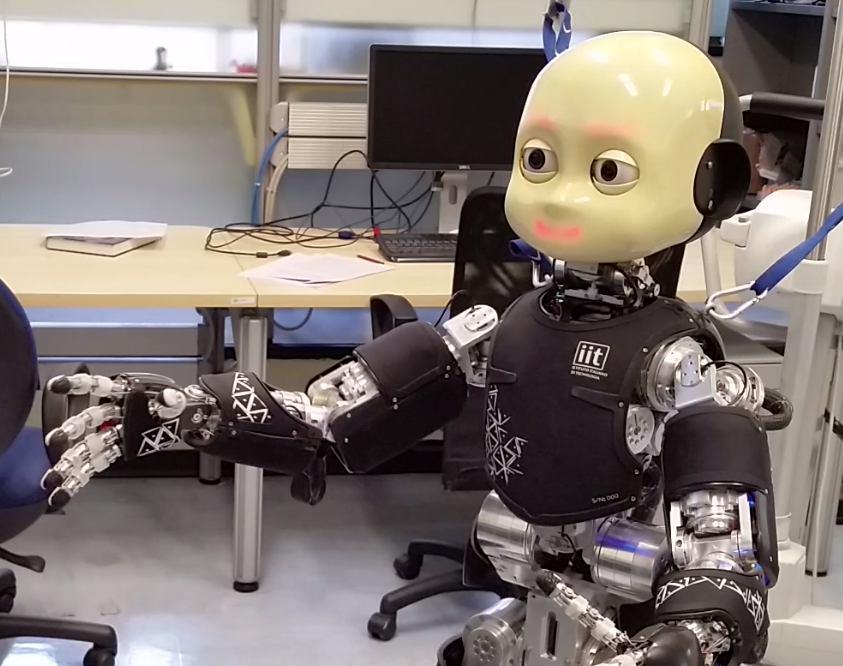}
\caption{iCub learning its right arm dynamics.\label{overflow}}
\end{figure}

The article is organized as follows. 
Section \ref{sec:background} introduces the existing techniques for parametric and nonparametric robot dynamics learning.
In Section \ref{sec:semiparametric}, a complete description of the proposed semiparametric incremental learning technique is introduced. 
Section \ref{sec:experiments} presents the validation of our approach on the iCub humanoid robotic platform.
 Finally, Section \ref{sec:conclusion} summarizes the content of our work.

\section{BACKGROUND}
\label{sec:background}

\subsection{Notation}
The following notation is used throughout the paper.
\begin{itemize}
 \item The set of real numbers is denoted by $\mathbb{R}$. Let $\mathbf{u}$ and $\mathbf{v}$ be two $n$-dimensional column vectors of real numbers (unless specified otherwise), i.e. $\mathbf{u},\mathbf{v} \in \mathbb{R}^n$, 
 their inner product is denoted as $\mathbf{u}^\top \mathbf{v}$, with ``$\top$'' the transpose operator.
 \item The Frobenius norm of either a vector or a matrix of real numbers is denoted by $\| \cdot \|$.
\item $I_n \in \mathbb{R}^{n \times n}$ denotes the identity matrix of dimension~$n$; 
$0_n \in \mathbb{R}^n$ denotes the zero column vector of dimension~$n$; $0_{n \times m} \in \mathbb{R}^{n \times m}$ denotes the zero matrix of dimension~$n \times m$.
\end{itemize}

\subsection{Parametric Models of Robot Dynamics}
\label{sec:parmod}
Robot dynamics parametric models are used to represent the relation
connecting the geometric and inertial parameters with some dynamic quantities 
that depend uniquely on the robot model.
A typical example is obtained by writing the robot inverse dynamics equation in linear form with respect to the 
robot inertial parameters $\boldsymbol\pi$:
\begin{equation}
\boldsymbol\tau = M(\mathbf{q})\mathbf{\ddot{q}} + C(\mathbf{q},\mathbf{\dot{q}})\mathbf{\dot{q}} + g(\mathbf{q}) = \regressor \boldsymbol\pi,
\end{equation}
where:
$\mathbf{q} \in \mathbb{R}^{n_{dof}}$ is the vector of joint positions,
$\boldsymbol\tau \in \mathbb{R}^{n_{dof}}$ is the vector of joint torques,
$\boldsymbol\pi  \in \mathbb{R}^{n_p}$ is the vector of the identifiable (base) inertial parameters \cite{reference/robo/FeatherstoneO08},
$\regressor \in \mathbb{R}^{n_{dof} \times n_p}$ is the ``regressor'', i.e. a matrix that depends only 
on the robot kinematic parameters. In the rest of the paper we will indicate with $\mathbf{x}$ the triple given by $(\mathbf{q},\mathbf{\dot{q}},\mathbf{\ddot{q}})$.
Other parametric models write different measurable quantities as a product 
of a regressor and a vector of parameters, for example  
the total energy of the robot \cite{gautier1988identification}, the istantaneous power provided to the robot \cite{gautier1997dynamic}, the sum of all external forces acting on the robot \cite{journal/ayusawa2014} or 
the center of pressure of the ground reaction forces \cite{conf/baelemans2013}. Regardless of the choice of the  measured variable $\mathbf{y}$, 
the structure of the regressor is similar:
\begin{equation}
\label{eq:parametricModel}
\mathbf{y} = \Phi(\mathbf{q},\mathbf{\dot{q}},\mathbf{\ddot{q}}) \boldsymbol\pi = \regressor \boldsymbol\pi ,
\end{equation}
where $\mathbf{y} \in \mathbb{R}^{t}$ is the measured quantity.

The $\inertialParameters$ vector is composed of certain linear combinations of the inertial parameters of the links, the \emph{base inertial parameters} \cite{khalil2004modeling}. 
In particular, the inertial parameters of a single body are the mass $m$, 
the first moment of mass $m\mathbf{c} \in \mathbb{R}^3$ expressed in a body fixed frame and the inertia matrix $I\in \mathbb{R}^{3 \times 3}$ expressed 
in the orientation of the body fixed frame and with respect to its origin.  

In \emph{parametric modeling} of robot dynamics, the regressor structure depends on the kinematic parameters 
of the robot, that are obtained from CAD models of the robot through kinematic calibration techniques. 
Similarly, the inertial parameters $\boldsymbol\pi$ can also be obtained from CAD models of the robot, 
however these models may be unavailable (for example) because the manufacturer of the robot does not provide them. 
In this case the usual approach is to estimate $\boldsymbol\pi$ from experimental data \cite{hollerbach2008model}.
To do that, given $\nrOfSamples$ measures of the measured quantity $\mathbf{y}_i$ (with $i = 1 \dots \nrOfSamples$), stacking \eqref{eq:parametricModel} for the
$\nrOfSamples$ samples it is possible to write:

\begin{equation}
 \label{eq:LSParametric}
 \begin{bmatrix}
 \outputQuant_1 \\
 \outputQuant_2 \\
 \vdots       \\
 \outputQuant_{\nrOfSamples} 
 \end{bmatrix}
 = 
 \begin{bmatrix}
 \regressorNum{1} \\
 \regressorNum{2} \\
 \vdots           \\
 \regressorNum{\nrOfSamples} 
 \end{bmatrix}
 \boldsymbol\pi .
\end{equation}

This equation can then be solved in least squares (LS) sense to find an estimate $\estimatedInertialParameters$ of the base inertial 
parameters. Given the training trajectories it is possible that not all parameters in $\inertialParameters$ can be estimated well as the problem in \eqref{eq:LSParametric}
can be ill-posed, hence this equation is usually solved as a \emph{Regularized Least Squares} (RLS) problem. Defining 
$$
\overline{\outputQuant}_{\nrOfSamples} = \begin{bmatrix}
 \outputQuant_1 \\
 \outputQuant_2 \\
 \vdots       \\
 \outputQuant_{\nrOfSamples} 
 \end{bmatrix} ,
 \hspace{1em}
 \overline{\bm{\Phi}}_{\nrOfSamples} = \begin{bmatrix}
 \regressorNum{1} \\
 \regressorNum{2} \\
 \vdots           \\
 \regressorNum{\nrOfSamples} 
 \end{bmatrix} ,$$
 the RLS problem that is solved for the parametric identification is:
\begin{equation}
 \label{eq:parametricRLS}
  \estimatedInertialParameters = \argmin\limits_{\inertialParameters \in  \mathbb{R}^{n_p}} \left( \|\overline{\bm{\Phi}}_{\nrOfSamples} \inertialParameters -\overline{\outputQuant}_{\nrOfSamples}\|^2 + \lambda \| \inertialParameters \|^2 \right) , \lambda > 0.
\end{equation}


\subsection{Nonparametric Modeling with Kernel Methods}
\label{sec:np_modeling}

Consider a probability distribution $\rho$ over the probability space $\mathcal{X} \times \mathcal{Y}$, where $\mathcal{X} \subseteq \mathbb{R}^d$ is the input space (the space of the $d$ measured attributes) and $\mathcal{Y}\subseteq \mathbb{R}^t$ is the output space (the space of the $t$ outputs to be predicted). In a nonparametric modeling setting, the goal is to find a function $f^*: \mathcal{X} \rightarrow \mathcal{Y}$ belonging to a set of measurable functions $\mathcal{H}$, called \textit{hypothesis space}, such that
\begin{equation}
\label{eq:rm}
f^* = \argmin\limits_{f \in \mathcal{H}} 
\underbrace{\int_{\mathcal{X} \times \mathcal{Y}} \ell(f(\mathbf{x}),\mathbf{y}) d\rho(\mathbf{x},\mathbf{y})}_{\mathcal{E}(f)},
\end{equation}
where $\mathbf{x} \in \mathcal{X}$ are row vectors, $\mathbf{y}\in \mathcal{Y}$, $\mathcal{E}(f)$ is called \textit{expected risk} and $\ell(f(\mathbf{x}),\mathbf{y})$ is the \textit{loss function}. In the rest of this work, we will consider the squared loss $\ell(f(\mathbf{x}),\mathbf{y}) = \|f(\mathbf{x})-\mathbf{y}\|^2$.
Note that the distribution $\rho$ is unknown, and that we assume to have access to a discrete and finite set of measured data points $S = \lbrace \mathbf{x}_i,\mathbf{y}_i\rbrace_{i=1}^n$ of cardinality $n$, in which the points are independently and identically distributed (i.i.d.) according to $\rho$.\\
In the context of kernel methods \cite{schlkopf2002learning}, $\mathcal{H}$ is a \textit{reproducing kernel Hilbert space} (RKHS). An RKHS is a Hilbert space of functions such that $\exists k : \mathcal{X} \times \mathcal{X} \rightarrow \mathbb{R}$ for which the following properties hold:
\begin{enumerate}
\item $\forall \mathbf{x} \in \mathcal{X} \quad k_\mathbf{x}(\cdot) = k(\mathbf{x},\cdot) \in \mathcal{H}$
\item $g(\mathbf{x}) = \left\langle  g, k_\mathbf{x}\right\rangle _{\mathcal{H}} \forall g \in \mathcal{H}, \mathbf{x} \in \mathcal{X}$ ,
\end{enumerate}
where $\left\langle  \cdot , \cdot \right\rangle _{\mathcal{H}} $ indicates the inner product in $\mathcal{H}$. The function $k$ is a \textit{reproducing kernel}, and it can be shown to be symmetric positive definite (SPD).
We also define  the kernel matrix $K \in \mathbb{R}^{n \times n} \quad s.t. \quad K_{i,j} = k(\mathbf{x}_i,\mathbf{x}_j)$, which is symmetric and positive semidefinite (SPSD) $\forall \mathbf{x}_{i}, \mathbf{x}_{j} \in \mathcal{X}$, with $i,j \in \lbrace 1, \ldots, n \rbrace, n \in \mathbb{N}^+$.

The optimization problem outlined in \eqref{eq:rm} can be approached empirically by means of many different algorithms, among which one of the most widely used is Kernel Regularized Least Squares (KRLS) \cite{SaundersGV98,rifkin2003regularized}.
In KRLS, a regularized solution $\hat{f}_{\lambda} : \mathcal{X} \rightarrow \mathcal{Y}$ is found solving
\begin{equation}
\label{eq:tik}
\hat{f}_{\lambda} = \argmin\limits_{f \in \mathcal{H}} \left( \sum_{i=1}^n \|f(\mathbf{x}_i) - \mathbf{y}_i\|^2 + \lambda \|f\|_{\mathcal{H}}^2 \right), \lambda > 0 ,
\end{equation}
where $\lambda$ is called \textit{regularization parameter}.
The solution  to \eqref{eq:tik} exists and is unique.
Following the representer theorem \cite{schlkopf2002learning}, the solution can be conveniently expressed as
\begin{equation}
\hat{f}_{\lambda}(\mathbf{x}) = \sum_{i = 1}^n \bm{\alpha}_i k(\mathbf{x}_i,\mathbf{x})
\end{equation}
with $\bm{\alpha} = (K+\lambda I_n)^{-1}Y \in \mathbb{R}^{n \times t}$, $\bm{\alpha}_i$ $i$-th row of $\bm{\alpha}$ and $Y = \left[ \mathbf{y}^\top_1, \ldots, \mathbf{y}^\top_n \right]^\top$.
It is therefore necessary to invert and store the kernel matrix $K \in \mathbb{R}^{n \times n}$, which implies $O(n^3)$ and $O(n^2)$ time and memory complexities, respectively.
Such complexities render the above-mentioned KRLS approach prohibitive in settings where $n$ is large, including the one treated in this work. This limitation can be dealt with by resorting to approximated methods such as \textit{random features}, which will now be described.

\subsubsection{Random Feature Maps for Kernel Approximation}
\label{sec:randfeats}
The random features approach was first introduced in \cite{RahimiR07}, and since then is has been widely applied in the field of large-scale Machine Learning. This approach leverages the fact that the kernel function can be expressed as 
\begin{equation}
\label{eq:innprod}
k(\mathbf{x},\mathbf{x}') = \left\langle \phi(\mathbf{x}) , \phi(\mathbf{x}') \right\rangle _{\mathcal{H}} ,
\end{equation}
where $\mathbf{x}, \mathbf{x}' \in \mathcal{X}$ are row vectors, $\phi: \mathbb{R}^d \rightarrow \mathbb{R}^p$ is a \textit{feature map} associated with the kernel, which maps the input points from the input space $\mathcal{X}$ to a feature space of dimensionality $p \leq + \infty$, depending on the chosen kernel.
When $p$ is very large, directly computing the inner product as in \eqref{eq:innprod} enables the computation of the solution, as we have seen for KRLS.
However, $K$ can become too cumbersome to invert and store as $n$ grows. A random feature map $\tilde{\phi}: \mathbb{R}^d \rightarrow \mathbb{R}^D$, typically with $D \ll p$, directly approximates the feature map $\phi$, so that
\begin{equation}
k(\mathbf{x},\mathbf{x}') = \left\langle \phi(\mathbf{x}) , \phi(\mathbf{x}') \right\rangle _{\mathcal{H}} \approx \tilde{\phi}(\mathbf{x})\tilde{\phi}(\mathbf{x}')^\top.
\end{equation}
$D$ can be chosen according to the desired approximation accuracy, as guaranteed by the convergence bounds reported in \cite{RahimiR07,4797607}.
In particular, we will use random Fourier features for approximating the Gaussian kernel
\begin{equation}
k(\mathbf{x},\mathbf{x}')=e^{-\frac{\|\mathbf{x}-\mathbf{x}'\|^2}{2\sigma^2}}.
\end{equation}
The approximated feature map in this case is $\tilde{\phi}(\mathbf{x}) = \left[ e^{i\mathbf{x} \bm{\omega}_1}, \ldots, e^{i\mathbf{x}\bm{\omega}_D} \right]$, where
\begin{equation}
\bm{\omega} \sim p(\bm{\omega}) = (2 \pi)^{-\frac{D}{2}}e^{-\frac{\|\bm{\omega}\|^2}{2\sigma^2}},
\end{equation}
with $\bm{\omega} \in \mathbb{R}^d$ column vector. The fundamental theoretical result on which random Fourier features approximation relies is Bochner's Theorem \cite{rudin1990fourier}.
The latter states that if $k(\mathbf{x},\mathbf{x}')$ is a shift-invariant kernel on $\mathbb{R}^d$, then $k$ is positive definite if and only if its Fourier transform $p(\omega)\geq0$.
If this holds, by the definition of Fourier transform we can write
\begin{equation}
k(\mathbf{x},\mathbf{x}') = k(\mathbf{x}-\mathbf{x}') = \int_{\mathbb{R}^d}p(\bm{\omega})e^{i(\mathbf{x}-\mathbf{x}')\bm{\omega}}d\bm{\omega} ,
\end{equation}
which can be approximated by performing an empirical average as follows:
\begin{equation}
\begin{array}{r@{}l}
k(\mathbf{x}-\mathbf{x}') &{}= \mathbb{E}_{\bm{\omega} \sim p} \left[e^{i(\mathbf{x}-\mathbf{x}')\bm{\omega}} \right] \approx \\
&{}\approx \frac{1}{D} \sum_{k=1}^De^{i(\mathbf{x}-\mathbf{x}')\bm{\omega}} = \tilde{\phi}(\mathbf{x}) \tilde{\phi}(\mathbf{x}')^\top.
\end{array}
\end{equation}
Therefore, it is possible to map the input data as $\tilde{\mathbf{x}} = \tilde{\phi}(\mathbf{x}) \in \mathbb{R}^{D}$, with $\tilde{\mathbf{x}}$ row vector, to obtain a nonlinear and nonparametric model of the form
\begin{equation}
\tilde{f}(\mathbf{x}) = \tilde{\mathbf{x}} \tilde{W} \approx \hat{f}_{\lambda}(\mathbf{x}) = \sum_{i = 1}^n \bm{\alpha}_i k(\mathbf{x}_i,\mathbf{x})
\end{equation}
approximating the exact kernelized solution $\hat{f}_{\lambda}(\mathbf{x})$, with $\tilde{W} \in \mathbb{R}^{D \times t}$. Note that the approximated model is nonlinear in the input space, but linear in the random features space. We can therefore introduce the regularized linear regression problem in the random features space as follows:
\begin{equation}
\label{eq:randfeatsminimizationprob}
\tilde{W}^\lambda = \argmin\limits_{\tilde{W} \in  \mathbb{R}^{d \times t}} \left( \| \tilde{X} \tilde{W} - Y \|^2 + \lambda \| \tilde{W} \|^2 \right) , \lambda > 0,
\end{equation}
where $\tilde{X} \in \mathbb{R}^{n \times D}$ is the matrix of the training inputs where each row has been mapped by $\tilde{\phi}$.
The main advantage of performing a random feature mapping is that it allows us to obtain a nonlinear model by applying linear regression methods.
For instance, Regularized Least Squares (RLS) can compute the solution $\tilde{W}^\lambda$ of \eqref{eq:randfeatsminimizationprob} with $O(nD^2)$ time and $O(D^2)$ memory complexities.
Once $\tilde{W}^\lambda$ is known, the prediction $\hat{\mathbf{y}} \in \mathbb{R}^{1 \times t}$ for a mapped sample $\tilde{\mathbf{x}} $ can be computed as $\hat{\mathbf{y}} = \tilde{\mathbf{x}} \tilde{W}^\lambda$.

\subsection{Regularized Least Squares}
%
%
Let $\RLSinputMat\in \mathbb{R}^{a\times b}$ and 
$\RLSoutputMat \in \mathbb{R}^{a\times c}$ be two matrices of real numbers, with $a,b,c \in \mathbb{N}^+$.
The Regularized Least Squares (RLS) algorithm computes a regularized solution $W^\lambda \in \mathbb{R}^{b\times c}$ 
of the potentially ill-posed problem $\RLSinputMat W = \RLSoutputMat $, enforcing its numerical stability. 
Considering the widely used Tikhonov regularization scheme, $W^\lambda \in \mathbb{R}^{b\times c}$ is the solution to the following problem:
\begin{equation}
\label{eq:rlsminimizationprob}
W^\lambda = \argmin\limits_{W \in  \mathbb{R}^{b \times c}} \underbrace{ \left( \| \RLSinputMat W - \RLSoutputMat \|^2 + \lambda \| W \|^2 \right)}_{J(W,\lambda)}  , \quad \lambda>0 ,
\end{equation}
where $\lambda$ is the regularization parameter. By taking the gradient of $J(W,\lambda)$ with respect to $W$ and equating it to zero, the minimizing solution can be written as
\begin{equation}
\label{eq:rlsSolution}
W^\lambda = (\RLSinputMat^\top \RLSinputMat + \lambda I_{b})^{-1}\RLSinputMat^\top \RLSoutputMat .
\end{equation}

Both the parametric identification problem \eqref{eq:parametricRLS} and the nonparametric random features problem \eqref{eq:randfeatsminimizationprob} are specific
instances of the general problem \eqref{eq:rlsminimizationprob}.

In particular, the parametric problem \eqref{eq:parametricRLS} is equivalent to \eqref{eq:rlsminimizationprob} with:
$$
W^\lambda = \hat{\boldsymbol\pi}, \hspace{1em} Z = \overline{\Phi}_n, \hspace{1em} U = \overline{\mathbf{y}}_n
$$
while the random features learning problem \eqref{eq:randfeatsminimizationprob} is equivalent to \eqref{eq:rlsminimizationprob} with:
$$
W^\lambda = \tilde{W}^\lambda , \hspace{1em} Z = \tilde{X} , \hspace{1em} U = Y.
$$
Hence, both problems for a given set of $n$ samples can be solved applying \eqref{eq:rlsSolution}.

\subsection{Recursive Regularized Least Squares (RRLS) with Cholesky Update}
\label{sec:recupdate}

In scenarios in which supervised samples become available sequentially, a very useful extension of the RLS algorithm consists in the definition of an update rule for the model which allows it to be incrementally trained, increasing adaptivity to changes of the system properties through time.
This algorithm is called Recursive Regularized Least Squares (RRLS).
We will consider RRLS with the Cholesky update rule \cite{bjoerck_least_squares96}, which is numerically more stable than others (e.g. the Sherman-Morrison-Woodbury update rule).
In adaptive filtering, this update rule is known as the \textit{QR algorithm} \cite{Sayed:2008:AF:1370975}.

Let us define $A = Z^\top Z + \lambda I_{b}$ with $\lambda > 0$ and $B = Z^\top U$. Our goal is to update the model (fully described by $A$ and $B$)
with a new supervised sample $(\mathbf{z}_{k+1}, \mathbf{u}_{k+1})$, with $\mathbf{z}_{k+1} \in \mathbb{R}^b$, $\mathbf{u}_{k+1}\in \mathbb{R}^c$ row vectors.

Consider the Cholesky decomposition $A = R^\top R$. It can always be obtained, since $A$ is positive definite for $\lambda > 0$.
Thus, we can express the update problem at step $k+1$ as:
\begin{equation}
\begin{array}{r@{}l}
A_{k+1} &{}= R^\top_{k+1} R_{k+1}\\
&{}= A_{k} + \RLSinputSample_{k+1}^\top \RLSinputSample_{k+1}\\
&{}= R^\top_{k} R_{k} + \RLSinputSample_{k+i}^\top \RLSinputSample_{k+1} , 
\end{array}
\end{equation}
where $R$ is full rank and unique, and $R_0 = \sqrt{\lambda} I_b$.\\
By defining
\begin{equation}
\tilde{R}_{k} =
\left[ \begin{array}{c}
R_{k} \\
 \RLSinputSample_{k+1} \end{array} \right] \in \mathbb{R}^{b+1 \times b},
\end{equation}
we can write $A_{k+1}=\tilde{R}_{k}^\top \tilde{R}_{k}$. However, in order to compute $R_{k+1}$ from the obtained $A_{k+1}$ it would
be necessary to recompute its Cholesky decomposition, requiring $O(b^3)$ computational time. There exists a procedure, based on Givens
rotations, which can be used to compute $R_{k+1}$ from $\tilde{R}_{k}$ with $O(b^2)$ time complexity.
A recursive expression can  be obtained also for $B_{k+1}$ as follows:

\begin{equation}
\label{eq:rrlsUpdate}
\begin{array}{r@{}l}
B_{k+1} &{}= \RLSinputMat_{k+1}^\top \RLSoutputMat_{k+1}\\
&{}= \RLSinputMat_{k}^\top \RLSoutputMat_{k} + \RLSinputSample_{k+1}^\top \RLSoutputSample_{k+1}.
\end{array}
\end{equation}
Once $R_{k+1}$ and $B_{k+1}$ are known, the updated weights matrix $W_k$ can be obtained via back and forward substitution as 
\begin{equation}
W_{k+1} = R_{k+1} \setminus (R^\top_{k+1} \setminus B_{k+1}) .
\end{equation}
The time complexity for updating $W$ is $O(b^2)$.

As for RLS, the RRLS incremental solution can be applied to both the parametric \eqref{eq:parametricRLS} and nonparametric  with random features \eqref{eq:randfeatsminimizationprob} problems, assuming $\lambda > 0$. 
In particular, RRLS can be applied to the parametric case by noting that the arrival of a new sample $\left( \Phi_r , \mathbf{y}_r \right)$ adds 
$t$ rows to $Z_k = \overline{\Phi}_{r-1}$ and $U_k = \overline{\mathbf{y}}_{r-1}$. Consequently, the update of $A$ must be decomposed 
in $t$ update steps using \eqref{eq:rrlsUpdate}. For each one of these $t$ steps we consider only one row of $\Phi_{r}$ and $\mathbf{y}^\top_r$, namely: 
$$
\RLSinputSample_{k+i} = (\Phi_r)_i , \hspace{1em} \RLSoutputSample_{k+i} = (\mathbf{y}^\top_r)_i , \hspace{1em} i = 1 \dots t 
$$
where $(V)_i$ is the $i$-th row of the matrix $V$.

For the nonparametric random features case, RRLS can be simply applied with:
$$
\RLSinputSample_{k+1} = \tilde{\mathbf{x}}_r , \hspace{1em} \RLSoutputSample_{k+1} = \mathbf{y}_r .
$$
where $\left(\tilde{\mathbf x}_r, \mathbf y_r \right)$ is the supervised sample which becomes available at step $r$. 

\section{SEMIPARAMETRIC INCREMENTAL DYNAMICS LEARNING}
\label{sec:semiparametric}

\begin{figure*}
\centering
\vspace{3mm}
\begin{overpic}[width=0.9\textwidth]{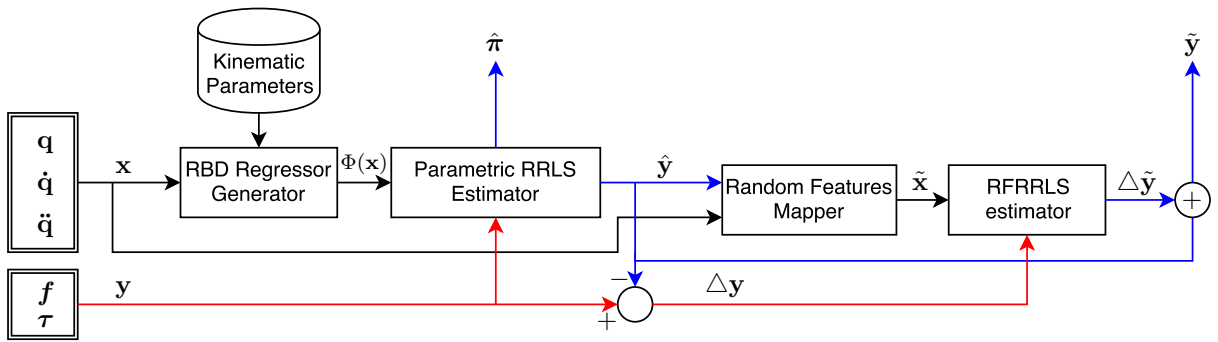}
\put (2.5,16) {$ \mathbf{q} $}
\put (2.5,12.5) {$ \mathbf{\dot{q}}  $}
\put (2.5,9) {$ \mathbf{\ddot{q}} $}
\put (2.5,3.3) {$ \bm{f} $}
\put (2.5,1.1) {$ \bm{\tau} $}
\put (9,13.9) {$ \mathbf{x} $}
\put (9,3.9) {$ \mathbf{y} $}
\put (27.7,14.2) {\footnotesize $ \Phi(\mathbf{x}) $}
\put (39.73,24) {$ \hat{\bm{\pi}}$}
\put (54,13.9) {$ \hat{\mathbf{y}} $}
\put (75.2,12.5) {$ \tilde{\mathbf{x}} $}
\put (92.2,12.5) {$ \triangle \tilde{\mathbf{y}} $}
\put (50,4.5) {$ - $}
\put (49,1) {$ + $}
\put (58,4) {$ \triangle \mathbf{y} $}
\put (97.6,11.1) {$ + $}
\put (97.8,24) {$ \tilde{\mathbf{y}}$}
\end{overpic}
\caption{Block diagram displaying the functioning of the proposed prioritized semiparametric inverse dynamics estimator. $ \bm{f} $ and $ \bm{\tau} $ indicate measured force and torque components, concatenated in the measured output vector $\mathbf{y}$. The parametric part is composed of the RBD regressor generator and of the parametric estimator based on RRLS. Its outputs are the estimated parameters $\hat{\bm{\pi}}$ and the predicted output $\hat{\mathbf{y}}$. The nonparametric part maps the input to the random features space with the Random Features Mapper block, and the RFRRLS estimator predicts the residual output $\triangle \tilde{\mathbf{y}} $, which is then added to the parametric prediction $\hat{\mathbf{y}}$ to obtain the semiparametric prediction $\tilde{\mathbf{y}}$.}
\label{fig:blockdia}
\end{figure*}

We propose a semiparametric incremental inverse dynamics estimator, designed to have better generalization properties with respect to fully parametric and nonparametric ones, both in terms of accuracy and convergence rates.
The estimator, whose functioning is illustrated by the block diagram in Figure \ref{fig:blockdia}, is composed of two main parts.
The first one is an incremental parametric estimator taking as input the rigid body dynamics regressors $ \Phi(x) $ and computing two quantities at each step:
\begin{itemize}
\item An estimate $\hat{\mathbf{y}}$ of the output quantities of interest
\item An estimate $\hat{\bm{\pi}}$ of the base inertial parameters of the links composing the rigid body structure
\end{itemize}
The employed learning algorithm is RRLS. Since it is supervised, during the model update step the measured output $\mathbf{y}$ is used by the learning algorithm as ground truth.
The parametric estimation is performed in the first place, and it is independent of the nonparametric part. 
This property is desirable in order to give priority to the identification of the inertial parameters $\bm{\pi}$. 
Moreover, being the estimator incremental, the estimated inertial parameters $\hat{\bm{\pi}}$ adapt to changes in the inertial properties of the links, which can occur if the end-effector is holding a heavy object. 
Still, this adaptation cannot address changes in nonlinear effects which do not respect the rigid body assumptions.\\
The second estimator is also RRLS-based, fully nonparametric and incremental. It leverages the approximation of the kernel function via random Fourier features, as outlined in Section  \ref{sec:randfeats}, to obtain a nonlinear model which can be updated incrementally with constant update complexity $O(D^2)$, where $D$ is the dimensionality of the random feature space (see Section \ref{sec:recupdate}).
This estimator receives as inputs the current vectorized $\mathbf{x}$ and $\hat{\mathbf{y}}$, normalized and mapped to the random features space approximating an infinite-dimensional feature space introduced by the Gaussian kernel.
The supervised output is the residual $ \triangle \mathbf{y} = \mathbf{y} - \hat{\mathbf{y}}$.
The nonparametric estimator provides as output the estimate $ \triangle \tilde{\mathbf{y}} $ of the residual, which is then added to $\hat{\mathbf{y}}$ to obtain the semiparametric estimate $ \tilde{\mathbf{y}}$.
Similarly to the parametric part, in the nonparametric one the estimator's internal nonlinear model can be updated during operation, which constitutes an advantage in the case in which the robot has to explore a previously unseen area of the state space, or when the mechanical conditions change (e.g. due to wear, tear or temperature shifts).

\section{EXPERIMENTAL RESULTS}
\label{sec:experiments}

\subsection{Software}
For implementing the proposed algorithm we used two existing 
open source libraries. 
For the RRLS learning part we used GURLS \cite{tacchetti2013gurls}, 
a regression and classification library based on the Regularized Least Squares (RLS) algorithm, available for 
Matlab and C++. 
For the computations of the regressors $\Phi(\mathbf{q},\mathbf{\dot{q}},\mathbf{\ddot{q}})$
we used iDynTree%
\footnote{\url{https://github.com/robotology/idyntree}}
, a C++ dynamics library designed for free floating robots. Using SWIG~\cite{beazley1996swig}, iDynTree supports 
calling its algorithms in several programming languages, such as Python, Lua and Matlab. 
For producing the presented results, we used the Matlab interfaces of iDynTree and GURLS. 

\subsection{Robotic Platform}
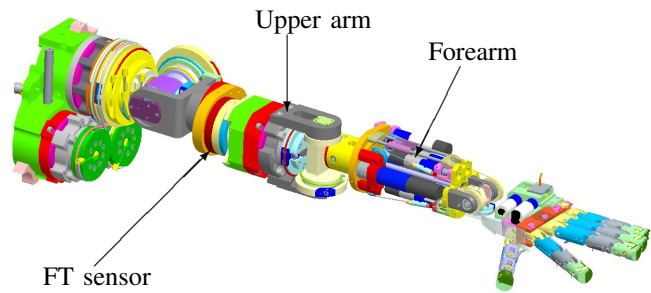
\begin{figure}[htb]
\begin{overpic}[width=0.48\textwidth,natwidth=1235,natheight=742]{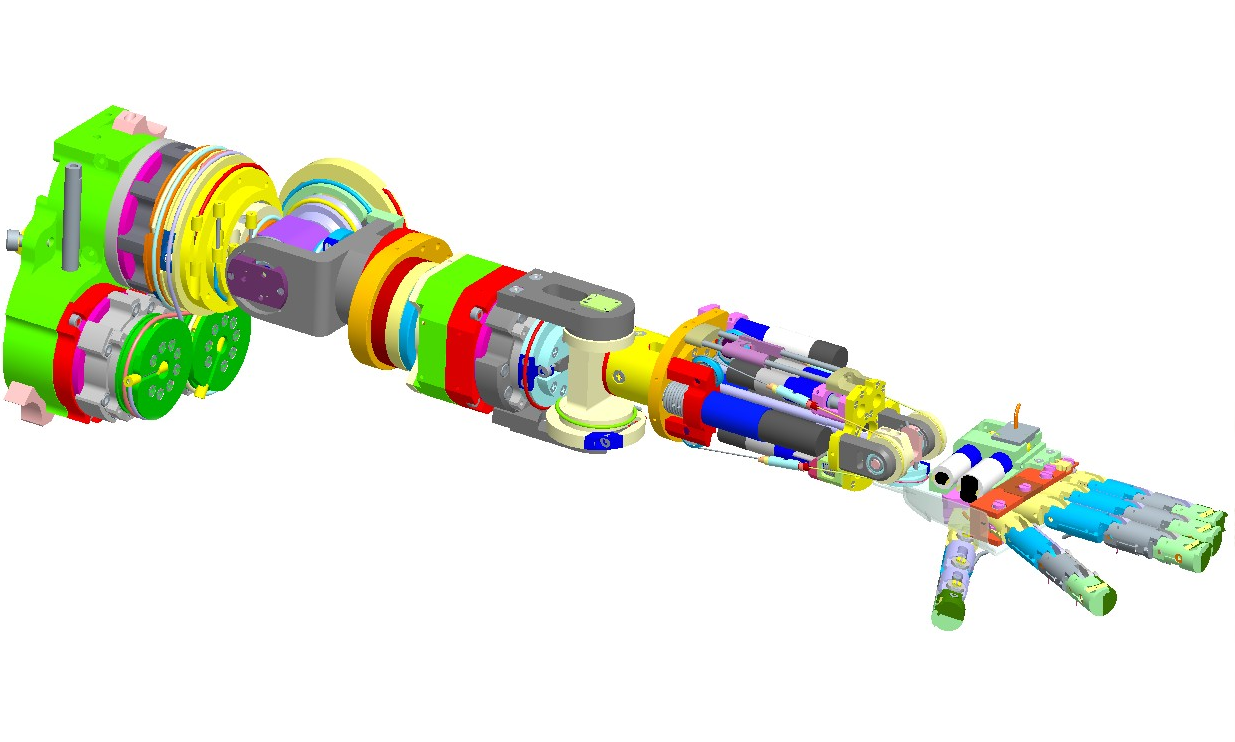}
\put(5,10){FT sensor}
\put(13,13){\vector(1,1){18}}
\put(38,50){Upper arm}
\put(43,49){\vector(0,-1){12}}
\put(65,45){Forearm}
\put(70,44){\vector(-1,-2){7}}
\end{overpic}
\caption{CAD drawing of the iCub arm used in the experiments. The six-axis F/T sensor used for validation is visible in the middle of the upper arm link.}
\label{fig:cadArm}
\end{figure}

iCub is a full-body humanoid with 53 degrees of freedom \cite{Metta2010}. For validating 
the presented approach, we learned the dynamics of the right arm of the iCub as measured
from the proximal six-axis force/torque (F/T) sensor embedded in the arm. 
The considered output $\mathbf{y}$ is the reading of the F/T sensor, and the inertial
parameters $\boldsymbol\pi$ are the base parameters of the arm~\cite{traversaro2015inertial}.
As $\mathbf{y}$ is not a input variable for the system, the output of the dynamic model is not directly usable for control, 
but it is still a proper benchmark for the dynamics learning problem, as also shown in ~\cite{GijsbertsM11}.
Nevertheless, the joint torques could be computed seamlessly from the F/T sensor readings  if needed for control purposes, by applying the method presented in \cite{6100813}.

\subsection{Validation}
\label{sec:validation}
\begin{figure}
\vspace{3mm}
\begin{subfigure}{.5\textwidth}
  \centering
  \includegraphics[width=0.93\linewidth]{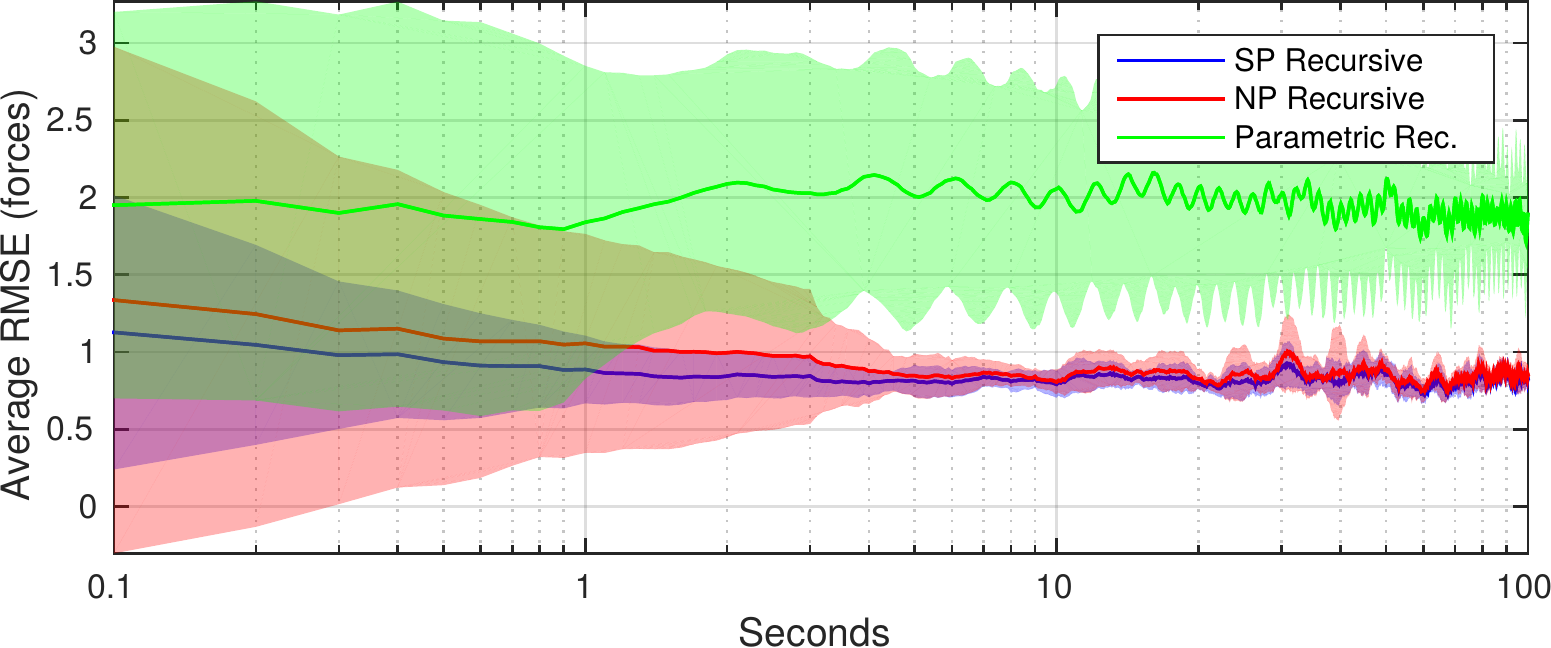}
  \vspace*{2mm}
\end{subfigure}
\begin{subfigure}{.5\textwidth}
  \centering
  \includegraphics[width=0.93\linewidth]{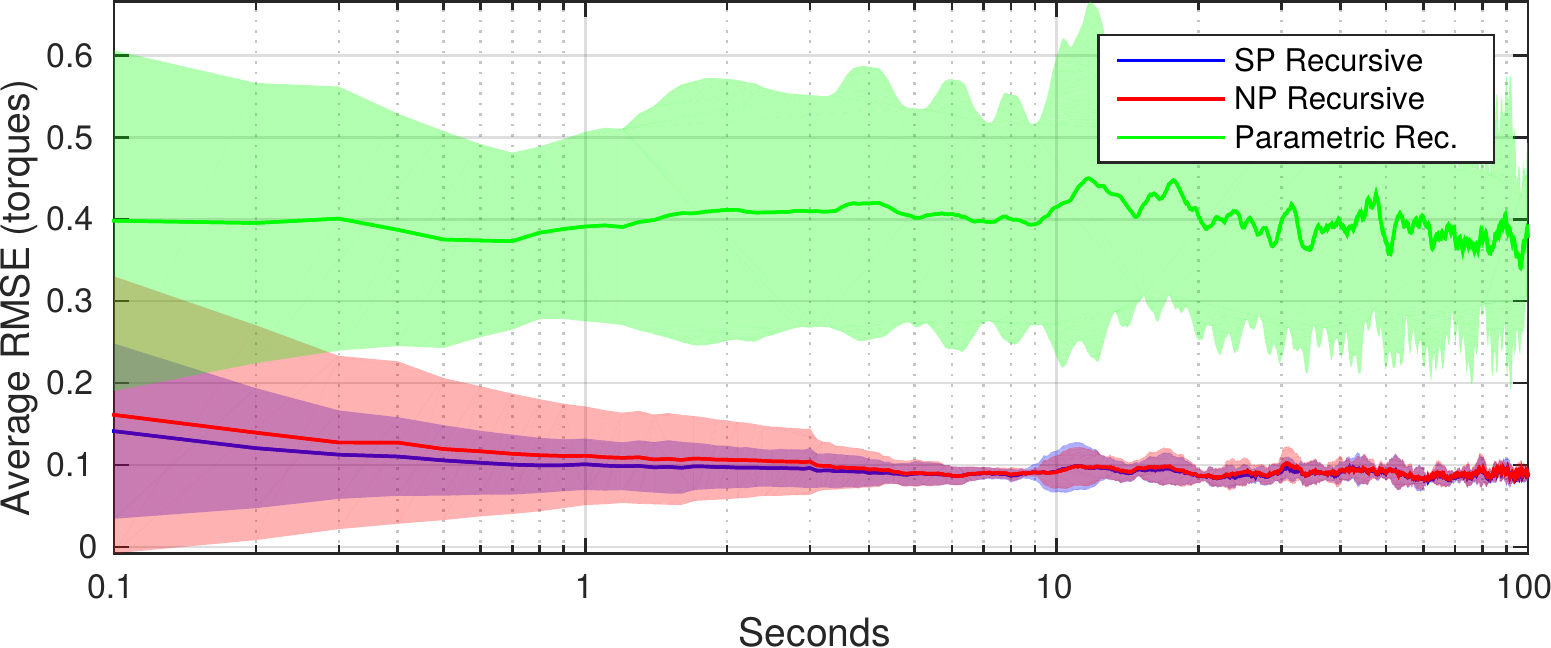}
\end{subfigure}
\caption{Predicted forces (top) and torques (bottom) components average RMSE, averaged over a 30-samples window for the recursive semiparametric (blue), nonparametric (red) and parametric (green) estimators. The solid lines indicate the mean values over 10 repetitions, and the transparent areas correspond to the standard deviations. On the $x$ axis, time (in seconds) is reported in logarithmic scale, in order to clearly show the behavior of the estimators during the initial transient phase. On the $y$ axis, the average RMSE is reported.}
\label{fig:exp1}
\end{figure}

The aim of this section is to present the results of the experimental validation of the proposed semiparametric model.
The model includes a parametric part which is based on physical modeling.
This part is expected to provide acceptable prediction accuracy for the force components in the whole workspace of the robot, since it is based on prior knowledge about the structure of the robot itself, which does not abruptly change as the trajectory changes.
On the other hand, the nonparametric part can provide higher prediction accuracy in specific areas of the input space for a given trajectory, since it also models nonrigid body dynamics effects by learning directly from data.
In order to provide empirical foundations to the above insights, a validation experiment has been set up using the right arm of the iCub humanoid robot, considering as input the positions, velocities and accelerations of the 3 shoulder joints and of the elbow joint, and as outputs the 3 force and 3 torque components measured by the six-axis F/T sensor in-built in the upper arm.
We employ two datasets for this experiment, collected at $10 Hz$ as the end-effector tracks (using the Cartesian controller presented in \cite{5650851}) circumferences with $10cm$ radius on the transverse ($XY$) and sagittal ($XZ$) planes\footnote{For more information on the iCub reference frames, see \url{http://eris.liralab.it/wiki/ICubForwardKinematics}} at approximately $0.6 m/s$. The total number of points for each dataset is $10000$, corresponding to approximately $17$ minutes of continuous operation.
The steps of the validation experiment for the three models are the following:
\begin{enumerate}
\item Initialize the recursive parametric, nonparametric and semiparametric models to zero. The inertial parameters are also initialized to zero
\item Train the models on the whole $XY$ dataset (10000 points)
\item Split the $XZ$ dataset in 10 sequential parts of 1000 samples each. Each part corresponds to 100 seconds of continuous operation
\item Test and update the models independently on the 10 splitted datasets, one sample at a time.
\end{enumerate}
In Figure \ref{fig:exp1} we present the means and standard deviations of the average root mean squared error (RMSE) of the predicted force and torque components on the 10 different test sets for the three models, averaged over a 3-seconds sliding window. 
The $x$ axis is reported in log-scale to facilitate the comparison of predictive performance for the different approaches in the initial transient phase.
We observe similar behaviors for the force and torque RMSEs. After few seconds, the nonparametric (NP) and semiparametric (SP) models provide more accurate predictions than the parametric (P) model with statistical significance.
At regime, their force prediction error is approximately $1N$, while the one of the P model is approximately two times larger. 
Similarly, the torque prediction error is $0.1Nm$ for SP and NP, which is considerably better than the $0.4Nm$ average RMSE of the P model.
It shall also be noted that the mean average RMSE of the SP model is lower than the NP one, both for forces and torques.
However, this slight difference is not very significant, since it is relatively small with respect to the standard deviation.
Given these experimental results, we can conclude that in terms of predictive accuracy the proposed incremental semiparametric method outperforms the incremental parametric one and matches the fully nonparametric one.
The SP method also shows a smaller standard deviation of the error with respect to the competing methods.
Considering the previous results and observations, the proposed method has been shown to be able to combine the main advantages of parametric modeling (i.e. interpretability) with the ones of nonparametric modeling (i.e. capacity of modeling nonrigid body dynamics phenomena).
The incremental nature of the algorithm, in both its P and NP parts, allows for adaptation to changing conditions of the robot itself and of the surrounding environment.

\section{CONCLUSIONS}
\label{sec:conclusion}

We presented a novel incremental semiparametric modeling approach for inverse dynamics learning, joining together the advantages of parametric modeling derived from rigid body dynamics equations and of nonparametric Machine Learning methods.
A distinctive trait of the proposed approach lies in its incremental nature, encompassing both the parametric and nonparametric parts and allowing for the prioritized update of both the identified base inertial parameters and the nonparametric weights.
Such feature is key to enabling robotic systems to adapt to mutable conditions of the environment and of their own mechanical properties throughout extended periods.
We validated our approach on the iCub humanoid robot, by analyzing the performances of a semiparametric inverse dynamics model of its right arm, comparing them with the ones obtained by state of the art fully nonparametric and parametric approaches.


\section*{ACKNOWLEDGMENT}
This paper was supported by the FP7 EU projects CoDyCo (No. 600716 ICT-2011.2.1 - Cognitive Systems and Robotics), Koroibot (No. 611909 ICT-2013.2.1 - Cognitive Systems and Robotics), WYSIWYD (No. 612139 ICT-2013.2.1 - Robotics, Cognitive Systems \& Smart Spaces, Symbiotic Interaction), and Xperience (No. 270273 ICT-2009.2.1 - Cognitive Systems and Robotics).


\bibliographystyle{IEEEtran}

\bibliography{icra2016}

\end{document}